\documentclass[final,5p,times,twocolumn]{elsarticle}
\usepackage{epsfig}
\usepackage{graphicx}
\usepackage{amsmath}
\usepackage{amssymb}
\usepackage{makecell,multirow}
\usepackage{booktabs} 
\usepackage{threeparttable}
\usepackage{graphicx} 
\usepackage{subfigure}
\usepackage{amsmath}
\usepackage{amssymb}
\usepackage{latexsym}
\usepackage{CJK}

\usepackage{tabularx}
\usepackage{array}

\usepackage{bigstrut}
\usepackage{bm}
\usepackage{graphicx} 
\usepackage{epstopdf}

\usepackage[linesnumbered,ruled,vlined]{algorithm2e}
\newsavebox\CBox
\def\textBF#1{\sbox\CBox{#1}\resizebox{\wd\CBox}{\ht\CBox}{\textbf{#1}}}

\usepackage[pagebackref=true,breaklinks=true,colorlinks,bookmarks=false]{hyperref}

\begin{document}
\begin{frontmatter}
\title{Stereo CenterNet based 3D Object Detection for Autonomous Driving}
\author[1,2,3]{Yuguang Shi}
\ead{syg@xs.ustb.edu.cn}
\author[1,4]{Yu Guo\corref{cor1}}
\ead{guoyu@ustb.edu.cn}
\author[1]{Zhenqiang Mi}
\ead{mizq@ustb.edu.cn}
\author[1]{Xinjie Li}
\ead{abcdvzz@hotmail.com}
\cortext[cor1]{Corresponding author}
\address[1]{School of Computer and Communication Engineering, University of Science and Technology Beijing, China}
\address[2]{Beijing Key Laboratory of Knowledge Engineering for Materials Science, University of Science and Technology Beijing, China}
\address[3]{Beijing Advanced Innovation Center for Materials Genome Engineering, University of Science and Technology Beijing, China}
\address[4]{Shunde Graduate School, University of Science and Technology, Guangdong, China}
\begin{abstract} Recently, three-dimensional (3D) detection based on stereo images has progressed remarkably; however, most advanced methods adopt anchor-based two-dimensional (2D) detection or depth estimation to address this problem. Nevertheless, high computational cost inhibits these methods from achieving real-time performance. In this study, we propose a 3D object detection method, Stereo CenterNet (SC), using geometric information in stereo imagery. SC predicts the four semantic key points of the 3D bounding box of the object in space and utilizes 2D left and right boxes, 3D dimension, orientation, and key points to restore the bounding box of the object in the 3D space. Subsequently, we adopt an improved photometric alignment module to further optimize the position of the 3D bounding box. Experiments conducted on the KITTI dataset indicate that the proposed SC exhibits the best speed-accuracy trade-off among advanced methods without using extra data.	
\end{abstract}
\end{frontmatter}

\section{Introduction}
Three-dimensional (3D) object detection is a fundamental but challenging task in several fields such as robotics and autonomous driving \cite{li2018stereo,cubeslam}. Several current mainstream 3D detectors rely on light detection and ranging (LiDAR) sensors to obtain accurate 3D information \cite{Pv-rcnn,3dssd,chen2017multi,Std}, and the application of LiDAR data has been considered crucial to the success of 3D detectors. Despite their substantial success and emerging low-cost LiDAR studies, it is important to note that LiDAR still faces a few challenges, such as its high cost, short life, and limited perception. Conversely, stereo cameras, which work in a manner resemble human binocular vision \cite{ZoomNet}, cost less and have higher resolutions; hence, they have garnered significant attention in academia and industry.

The basic theoretical knowledge of 3D detection based on stereo images can be traced back to triangulation \cite{stereo-matching} and the perspective n-point problem (pnp) \cite{epnp}. Owing to the introduction of cumbersome datasets \cite{ObjectNet3D,geiger2012we,ApolloScape,nuScenes}, 3D pose estimation entered the era of object detection. To date, machine learning-based methods have been widely adopted in practical engineering \cite{cubeslam,SVBS,Structured}. However, these methods have limited the ability to search for information in 3D space without requiring additional information; hence, it is difficult for its accuracy to exceed that of deep learning-based methods.

\begin{figure*}[t]
	\centering
	\includegraphics[width=14.9cm,height=6.5cm]{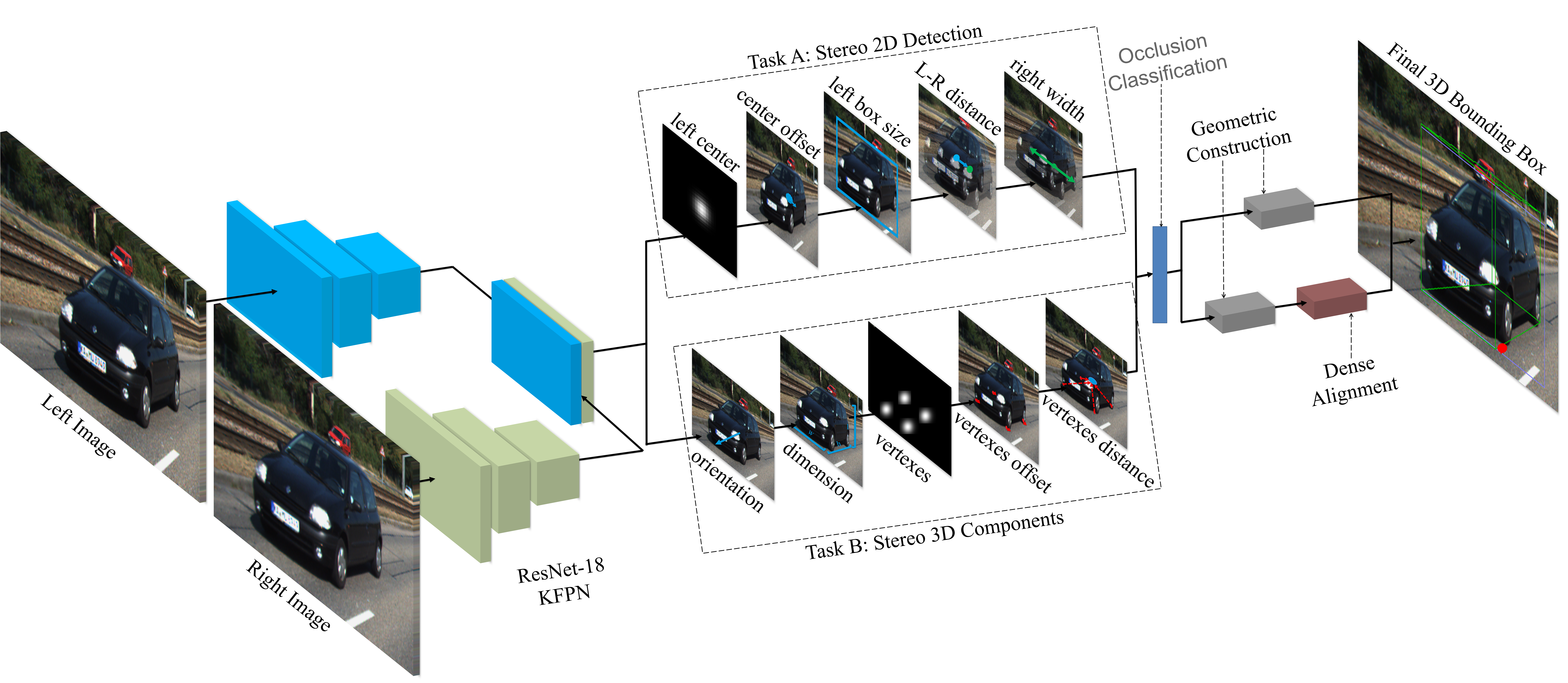}
	\caption{Network architecture of the proposed Stereo CenterNet that outputs 10 sub-branches for two tasks and the estimated 3D bounding box.}
	\label{b}
\end{figure*}

Recently, a few high-precision 3D detection methods for stereo images have emerged \cite{Stereor-cnn,wang2019pseudo,IDA,Dsgn} and have divided the 3D detection task into two subtasks: depth prediction and two-dimensional (2D) object detection. Regarding the depth prediction subtask, a number of methods adopt high-performance disparity estimation networks to calculate disparity map of the entire image. Other methods apply instance segmentation to solely predict disparity for pixels on objects of interest. However, a stereo 3D detector, e.g., Disp R-CNN \cite{Disp}, has a disparity estimation network that takes more than one third of the entire detection process time. Regarding the 2D object detection subtask, it supports the conditions in which the objects in KITTI images are typically small and most of them, which are heavily occluded, limit the performance of most present bottom-up detectors \cite{3DOP}. To ensure accuracy, almost all stereo-based 3D detectors rely on anchor-based 2D detectors and the association approach. 

However, anchor-based detectors and association approaches are faced by three primary limitations: $1)$ Hyperparameters of the anchor boxes, such as its size, aspect ratio, and number required to be carefully designed, that not only require relevant prior knowledge, but also influences the performance of the 2D detector. The existing method assigns the union of left and right GT boxes as the target for objectness classification. However, no research has been conducted on the parameters associated with stereo anchor boxes, and existing methods simply follow the parameters associated with monocular 2D detection. $2)$ Stereo anchor boxes incur additional processing steps and computation, such as scaling the stereo region of interest (ROI) using stereo ROI Align \cite{maskrcnn}, utilizing non-maximal suppression (NMS) to suppress overlapping 2D bounding boxes, and adopting real bounding boxes \cite{tian2019fcos} to compute the intersection of sets (IoU) scores. $3)$ Anchor-based detectors position the anchor boxes densely on the input image; however, only a part of the anchor boxes contain objects, triggering an imbalance between positive and negative samples, which does not only slow down the training, but may also lead to degenerate models. These factors limit the upper bound of the 3D object detection speed, which makes it difficult for existing stereo-based 3D object detection methods to achieve real-time performance.

Considering the above challenges, this study achieves the main task of stereo-based 3D detectors that are difficult to inference in real-time. We apply the semantic and geometric information in the stereo images to propose an efficient 3D object detection method that combines deep learning and geometry named Stereo CenterNet (SC), as presented in Fig. \ref{b}. This method adopts an anchor-free single-stage structure. It solely adopts stereo RGB images as input, and does not rely on pixel-level mask annotations and LiDAR data as supervision. We propose a novel stereo objects association method after experimenting with various combinations to address the problem that existing stereo images 3D detectors are difficult to achieve in actual time. The method circumvents redundant information as much as possible, solely adds two detection heads, returns the information required for object association more directly, eliminates the anchor box constraint, simplifies the overall framework, and improves the general operational efficiency. In addition, we optimize the perspective key points classification strategy to improve the accuracy without network classification and reduce the computational effort of the framework. Finally, in our experiments, we inferred that the conventional dense alignment approach does not work optimally in the case of heavily occluded and truncated objects. We deduce that the occluded objects have negligible amounts of effective image information, which is not sufficient for optimizing the depth information. Therefore, we redesigned the objects screening strategy of the conventional dense alignment module that adaptively selects different 3D box estimation methods according to the object occlusion level, to further improve the detection accuracy of hard samples. Owing to these ideas, the proposed SC can achieve a faster inference time while ensuring accuracy.

Our contributions are summarized as follows:
\begin{itemize}
	\item[$\bullet$]An anchor-free 2D box association method. To reduce calculation, we solely detect objects in the left image and perform left-right association by predicting a left-right distance.
	\item[$\bullet$]An adaptive object screening strategy, which selects different 3D box alignment methods according to the occlusion level of the object.
	\item[$\bullet$]Evaluation on the KITTI dataset. We propose a novel stereo-based 3D object detection method that does not require depth estimation and anchor boxes. We achieved a better speed-accuracy trade-off than other methods without using extra data.
\end{itemize}

\section{Related Work}

In this section, we briefly review recently presented literature on 3D object detection using monocular and stereo images.

\noindent{\bf Monocular image based methods:} GS3D \cite{li2019gs3d} extract features from the visible surface of the 3D bounding box to solve the feature blur problem. Shi Xuepeng et al. \cite{shi2020distance} adopted scale information to formulate different detection tasks, and also proposed a fully convolutional cascade point regression method that restores the spatial size and direction of the object via the loss of projection consistency. RTM3D \cite{RTM3D} predicts the nine perspective key points of the 3D bounding box of the image space, and adopts geometric relationships to restore the information of the object in the 3D space. MonoPair \cite{chen2020monopair} considered the relationship between paired samples to improve monocular 3D object detection. SMOKE \cite{SMOKE} detects the 3D center point of the object, and proposes a multi-step separation method to construct a 3D bounding box. According to Beker et al. \cite{beker2020monocular}, the 3D position and grid of each object in the image can be self-supervised using shape priors and instance masks. However, the monocular-image-based method finds it difficult to obtain accurate depth information.

\noindent{\bf Stereo images based methods:} According to the type of training data, stereo-imagery-based methods can be generally divided into three types. The first type solely requires stereo images with corresponding annotated 3D bounding boxes. TLNet \cite{qin2019triangulation} enumerates several 3D anchors between the ROI in stereo images to construct object-level correspondences, and introduces a channel reweighting strategy that facilitates the learning process. Stereo R-CNN \cite{Stereor-cnn} converted the 3D object detection problem to left and right 2D box, key point detection, dense alignment, including other tasks, and adopted geometric relationship generation constraints to develop a 3D bounding box; however, it did not solve the detection problem of occluded objects. IDA-3D \cite{IDA} predicts the depth of the object midpoint via the instance depth perception, parallax adaptation, and matching cost weighting methods, which can detect the 3D box end-to-end. Although these methods attempt to harness the potential of stereo images, its accuracy is the lowest among the three training types. The second type requires an additional depth map to train data. These methods \cite{wang2019pseudo,you2019pseudo,qian2020end} convert the estimated disparity map of the stereo images into pseudo LiDAR points, and adopt the LiDAR input method to estimate the 3D bounding box. DSGN \cite{Dsgn} re-encodes 3D objects and can detect 3D objects end-to-end; however, although it exhibits the highest performance among the three methods, it has a slow inference speed. Another one adds the instance segmentation annotation on the basis of the Pseudo-LiDAR training data. To minimize the amount of calculation and eliminate the disparity estimation network triggering streaking artifacts, these methods \cite{Disp,pon2020object,ZoomNet} adopt the stereo instance segmentation network to extract the pixels on the object of interest and predict disparity. It is evident that more information will definitely facilitate higher detection performance. However, the pixel-level mask annotations are significantly heavier than the frame-level annotations required for object detection, and the high training cost hinders the deployment of stereo systems in practical applications.

\section{Approach}

The overall network, which is built on CenterNet \cite{centernet}, uses a weight-share backbone network to extract consistent features on left and right images architecture, as illustrated in Fig. \ref{b}. The network outputs $10$ sub-branches behind the backbone network to complete two tasks: (A) stereo 2D detection and (B) stereo 3D Components. In task A, we used five sub-branches to estimate the left objects center, left objects center offset, left objects bounding box size, left and right center distances, and the width of the right objects box. The five sub-branches in task B estimated the orientation, dimension, bottom vertices, bottom vertices offset, vertices, and left objects center distance of the 3D bounding box. Subsequently, the objects in the image were dichotomized according to the level of occlusion, based on this information. Different methods are adopted to obtain the coordinates of objects in 3D space for heavily occluded objects and those that are not heavily occluded.
\label{SC}
\begin{figure}[t]
	\begin{center}
		\includegraphics[width=8cm,height=3.29cm]{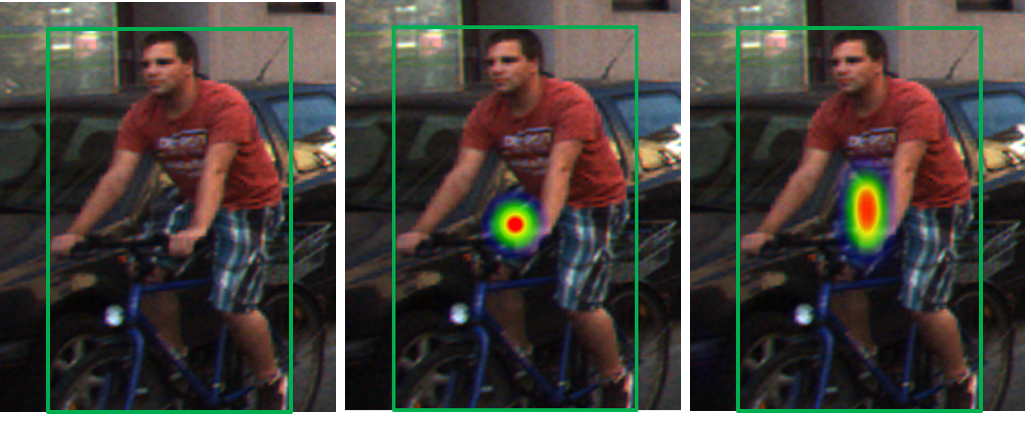}
	\end{center}
	\caption{Center point heatmaps generated via different gaussian kernels. From left to right: original image, heatmap without aspect ratio, and heatmap with aspect ratio.}
	\label{c}
\end{figure}

\subsection{Stereo CenterNet}

In this section, the backbone network of feature extraction is briefly discussed, and then the 2D and 3D detection modules are comprehensively introduced. Finally, the specific details of the implementation are introduced.

\noindent{\bf Backbone.} For testing, we adopted two different backbone networks: ResNet-18 \cite{resnet} and DLA-34 \cite{dla}. The two RGB images $i\in r^{2\times w_{i} \times h_{i} \times 3}$ of the stereo cameras were input into the model, and the down-sampled feature map was obtained four times, where $w_{i}$ and $h_{i}$ represent the width and height of the input image, respectively. Regarding ResNet-18, we adopted the keypoint feature pyramid network (KFPN) structure in RTM3D \cite{RTM3D} to increase the key point feature extraction capability. For DLA-34, the same structure used in CenterNet \cite{centernet} was adopted, and all hierarchical aggregation connections were replaced with deformable convnets networks (DCN) \cite{dcnv2}. We connected the stereo feature maps and added a 1$\times$1 convolutional layer to reduce the channel size. The number of channels output by ResNet-18 and DLA-34 were 128 and 256, respectively. Each of the $10$ output sub-branches was connected to the backbone network with two convolutions of sizes $3\times 3\times \left ( 128/256 \right )$ and $1\times 1\times n$, where $n$ denotes the characteristic channel of the relevant output branch. All $10$ sub-branches maintained the same characteristic width and height.

\noindent{\bf Stereo 2D Detection.} CenterNet \cite{centernet} is a bottom-up anchor-free object detection method. The 2D bounding box of the object was constructed by predicting the center point of the object and the object's width and height. CenterNet has three regression branches: center point coordinates, 2D size, and center offset.
Similar to CenterNet, we adopted the center point of the left-image object as the main center connecting all functions, and by introducing new branches to detect and associate the 2D bounding boxes of the left and right images simultaneously, our stereo 2D detection has five branches: $\left [\left ( u_{l}, v_{l} \right ) , \left ( w_{l}, h_{l} \right ), off_{l}, dis, w_{r}\right ] $, where $\left ( u_{l}, v_{l} \right )$ represent the coordinates of the center point of the left image 2D box; in addition, $( w_{l}, h_{l})$, $off_{l}$, $dis$, and $w_{r}$ represent the width and height of the left 2D bounding box, local offset of the left image center point, distance between the right and left center points, and width of the 2D box on the right image, respectively. We adopted the corrected stereo images, and the left and right objects shared the height of the same 2D box; hence, it was not necessary to predict the right box height.

\begin{figure}[t]
	\begin{center}
		\includegraphics[width=8cm,height=3.295cm]{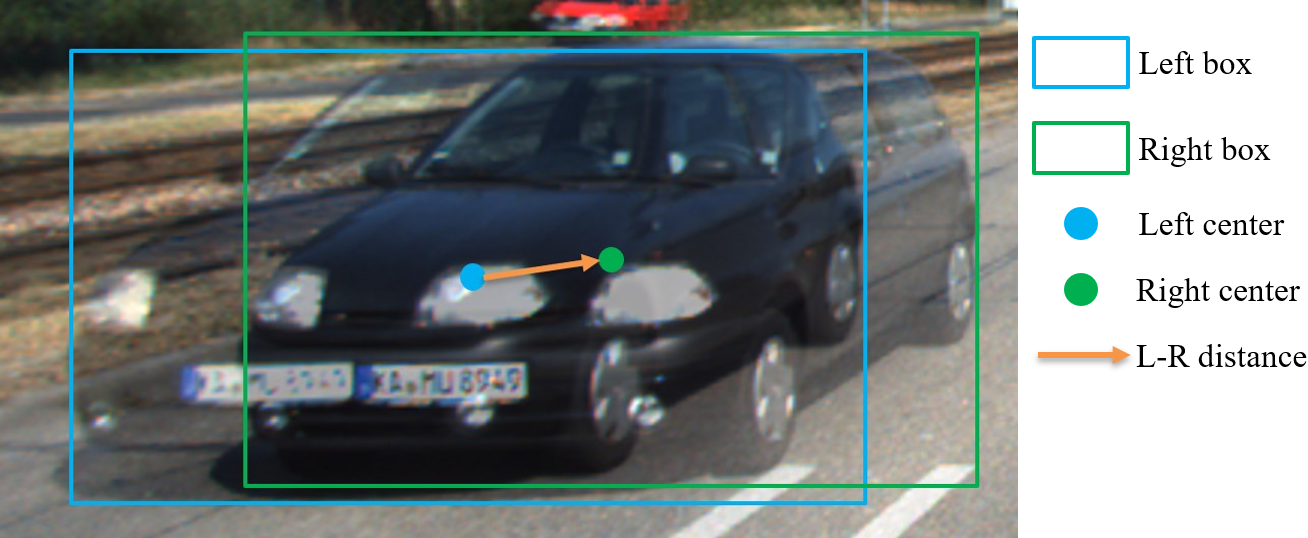}
	\end{center}
	\caption{Left and right objects are related by distance.}
	\label{d}
\end{figure}

The difference is that we chose a Gaussian kernel with aspect ratio \cite{ttf} to replace the Gaussian kernel in CenterNet \cite{centernet}, as illustrated in figure \ref{c}. The new heat map agrees more with people's intuitive judgments of objects, and can also provide more positive samples that can improve the detection accuracy of small objects. Regarding the main center point of the left object, we adopted a 2D Gaussian kernel $Y_{uvc} =exp\left ( - \frac{\left ( u-u_{l} \right )^{ 2} }{2\sigma _{x}^{2} }- \frac{\left (v-v_{l} \right )^{ 2} }{2\sigma _{y}^{2} }\right )$ to generate a heat map $Y\in\left [0,1 \right ] ^{\frac{w_{i} }{R}\times \frac{h_{i} }{R} \times C\times N }$, where $\sigma _{x}=\frac{\alpha w_{l}}{6}$,$\sigma _{y}=\frac{\alpha h_{l}}{6} $, $\alpha$ is an object size adaptive standard deviation \cite{law2018cornernet}. $C$, $R$, and $N$ denote the number of categories, downsampling multiple, and bitch size, respectively. Generate a heat map of key points during the prediction $\hat{Y} \in \left [ 0,1 \right ] ^{\frac{w_{i} }{R}\times \frac{h_{i} }{R} \times C\times N }$, and consider the predicted value $\hat{Y}_{xyc}=1$ as a positive sample, and $\hat{Y}_{xyc}=0$ as a negative sample. To address the imbalance problem of positive and negative samples, we adopted focal loss \cite{focalloss} to train.
\begin{eqnarray}
L_{m}= \frac{-1}{N} \sum_{uvc}^{} 
\begin{cases}
\left(1-\hat{Y}_{uvc}\right )^{\alpha }\log_{}{\left ( \hat{Y}_{uvc} \right ) } &ifY_{uvc}=1 \\\left ( 1-Y_{uvc}   \right )^{\beta }\left ( \hat{Y}_{uvc} \right )^{\alpha } 
\\ \ \  \ \log_{} {\left ( 1-\hat{Y}_{uvc} \right ) }&otherwise
\end{cases}
\end{eqnarray}
Where $N$ represents the number of key points in the image, while $\alpha$ and $\beta$ are hyperparameters set to $2$ and $4$, respectively, in the experiment.

We regressed the local offset of the left center point $\hat{F} \in \Re^{\frac{w_{i} }{R}\times \frac{h_{i} }{R} \times 2 }$ for each left ground truth center point $\left ( u_{l}, v_{l} \right)\in \Re ^{2}$, and the offset and distance were trained with an L1 loss.
\begin{eqnarray}
L_{off} = \frac{1}{N} \sum_{\left ( u_{l}, v_{l} \right)}^{} \left | \hat{F}_{\tilde{ \left ( u_{l}, v_{l} \right)}} -\left (\frac{ \left ( u_{l}, v_{l} \right)}{R} -\tilde{ \left ( u_{l}, v_{l} \right)}  \right )   \right | 
\end{eqnarray}
Where $\tilde{ \left ( u_{l}, v_{l} \right)}$ denotes the left center point, which was downsampled and then rounded to the integer value. Next, we used L1 loss to predict a left 2D bounding box size $S_{k}= ( w_{l}^{\left ( k \right ) } , h_{l}^{\left ( k \right ) })$ for each left object $k$. We adopted a single size prediction $\hat{S} \in\Re ^{\frac{w_{i} }{R}\times \frac{h_{i} }{R} \times 2 }$ for all object categories.
\begin{align}
L_{size} = \frac{1}{N} \sum_{k=1}^{N} \left | \hat{S}_{k} -S_{k} \right | 
\end{align}

To detect the right object, a feasible approach is to directly regress multiple identical detection heads after the left object detection. However, we decided to adopt another method to solely regress the information required to construct the 3D box, i.e., solely the left-right distance and width of the right object. 
Owing to perspective, the distance between the center of the left and right objects appears closer with the increase in depth. The primary motivation behind our approach is the idea that the left-right distance can be considered a special case of the left object center offset, thereby circumventing the increased computational effort of regressing other detection heads to obtain the right object center. In the ablation study (sect. \ref{Ablation}) we provide experimental results to prove the validity of this idea. Hence, we regressed the distance $\hat{D}\in\Re ^{\frac{w_{i} }{R}\times \frac{h_{i} }{R} \times 2 }$ between the left and right object center when the ground truth center $\left( u_{r}, v_{r} \right)\in \Re ^{2}$ of the right image object was known.
\begin{eqnarray}
L_{dis} = \frac{1}{N} \sum_{\left ( u_{r}, v_{r} \right)}^{} \left | \hat{D}_{\left ( u_{r}, v_{r} \right)} -\left (\frac{ \left ( u_{r}, v_{r} \right)}{R} -\tilde{ \left ( u_{l}, v_{l} \right)}  \right )   \right | 
\end{eqnarray}
To return to the width of the right-image object, similar to \cite{eigen2014depth}, we added an output transformation $w_{r}=1/\sigma \left ( \hat{w_{r}} \right ) -1$ before L1 loss, where $\sigma$ denotes the sigmoid function.
\begin{align}
L_{W_{r} } = \frac{1}{N} \sum_{k=1}^{N} \left | \frac{1}{\sigma \left ( \hat{w}_{r} \right ) }-1  -w_{r} \right | 
\end{align}
For the inference, use the $3\times3$ max pooling operation instead of NMS.

\noindent{\bf Stereo 3D Components} To build a 3D bounding box, we added three regression branches $\left [ d,o,v\right ]$ , where $d $ denotes the length, the width and height of the 3D box of the object $\left [L,W,H\right ]^{\top }$, $o$ represent the orientation of each object, and $v$ denotes the bottom vertex of the 3D box as the key point. Although the constraints of these three branches can restore the 3D information of the object, the impact of key points on the result is crucial. To further improve the accuracy of the results, two additional optional branches are provided: bottom vertex offset and main center point-vertex distance.

Regarding the 3D dimensions of the object, we pre-calculated the average of the 3D dimensions of a category in the entire data set $\left [\bar{L} ,\bar{W},\bar{H}\right ]^{\top}$, regressing to the true value and the offset between the prior size $off_{dim} = 2\times \left ( \left [L,W,H\right ] ^{\top }-\left [\bar{L} ,\bar{W},\bar{H}\right ] ^{\top }\right ) $. We trained the dimensions estimator using an L1 loss , and adopted the predicted size offset to restore the size of each object during inference.
\begin{align}
L_{dim} = \frac{1}{N} \sum_{k=1}^{N} \left | \hat{off}_{dim}^{k} -off_{dim}^{k} \right |
\end{align}
\begin{figure}[t]
	\begin{center}
		\includegraphics[width=5cm,height=5cm]{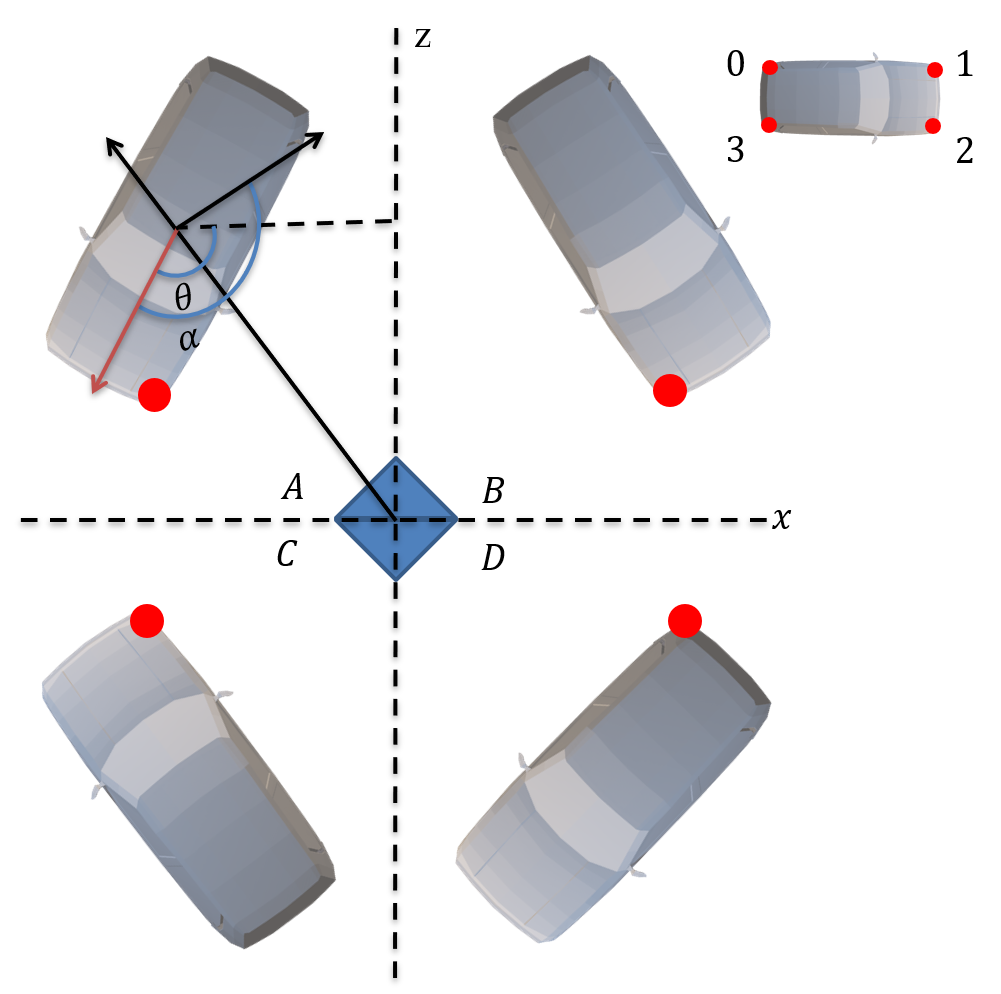}
	\end{center}
	\caption{Aerial view, the same car in different perspectives: A, B, C, D.}
	\label{e}
\end{figure}

For orientation, as illustrated in Figure \ref {e}, we regressed the car's local direction $\alpha$ instead of yaw rotation $\theta$. Similar to \cite{mousavian20173d}, we generated a feature map that uses eight scalars to indicate orientation $O\in\Re ^{\frac{w_{i} }{R}\times \frac{h_{i} }{R} \times 8 }$, the orientation were trained with the L1 loss. Subsequently, we utilized $\alpha$ and the object position to restore the yaw angle $\theta $.
\begin{align}
\theta =\alpha +arctan\left ( \frac{x}{z}\right )
\end{align}

To construct more stringent constraints, we predicted the four vertices at the bottom of the 3D bounding box, following \cite{Stereor-cnn}, and solely performed keypoint detection on the left image. We adopted the Gaussian kernel $V_{xyv}=exp\left ( - \frac{\left (x -x_{v} \right )^{ 2}+ \left (y -y_{v} \right )^{ 2}}{2\sigma _{v}^{2} } \right )$ to generate a ground truth vertex heat map $V\in \left [ 0,1 \right ] ^{\frac{w_{i} }{R}\times \frac{h_{i} }{R} \times 4}$, which is the same as the main center point of the left image, trained by focal loss. To improve the accuracy of the key points, we regressed the downsampling offset $F_{v}\in \Re^{\frac{w_{i} }{R}\times \frac{h_{i} }{R} \times2 }$ of each vertex.

To correlate the vertices with the center of the left image, we also returned the distance $D_{v} \in \Re^{\frac{w_{i} }{R}\times \frac{h_{i} }{R} \times 8 }$ from the main center to each vertex, and both the vertex offset and vertex distance applied the L1 loss.
We define the total loss of multitasking as:
\begin{eqnarray}
\begin{split}
L= &\omega_{m}L_{m}+\omega_{off} L_{off}+\omega_{dis}L_{dis}+\omega_{size}L_{size}\\ &+\omega_{W_{r}}L_{W_{r}}+
\omega_{dim}L_{dim}+\omega_{o}L_{o}+L_{v}L_{v}\\ &+ \omega_{off_{v} }L_{off_{v} }+\omega_{dis_{v}}L_{dis_{v}}
\end{split}
\end{eqnarray}

where $L_{o}, L_{v}, L_{off_{v} },$ and $L_{dis_{v}}$ represents orientation, vertex coordinate, vertex coordinate offset, and vertex coordinate distance, respectively. For the parameter $\omega$ before each item, we adopted uncertainty weighting\cite{kendall2018multi} instead of manual tuning
\label{sosp}

\subsection{Stereo 3D Box Estimation}
\noindent{\bf Geometric construction} After predicting the left and right 2D boxes, key points, angles, and 3D dimensions, following \cite{Stereor-cnn}, the center position and horizontal direction $K=\left \{x,y,z,\theta\right \} $ of the 3D box can be solved by minimizing the reprojection error of the 2D box and key points.

As presented in the figure \ref {b}, among the four bottom vertices, they can be accurately projected to the middle of the 2D bounding box, and the bottom vertices used to construct the 3D box are called perspective keypoints \cite{Stereor-cnn}. To identify the perspective key points in the four bottom vertices, we adopted logical judgment instead of image classification. As illustrated in Figure \ref{e}, we defined these four vertices as 0, 1, 2, 3, respectively, and the key points of perspective observed in different perspectives of the same car were different. However, in the same perspective, the key points of the perspective observed under the same car angle were the same. We deduce that the perspective key points are the same as the bottom vertex closest to the camera; hence, the perspective key points can be distinguished from the four vertices based on the predicted angle to estimate the 3D bounding box. The two key points of the boundary will be used for dense alignment. This method enables us to detect the keypoint of occluded and truncated objects, and improve the accuracy of hard samples.

\begin{figure}[t]
	\begin{center}
		\includegraphics[width=6.767cm,height=5cm]{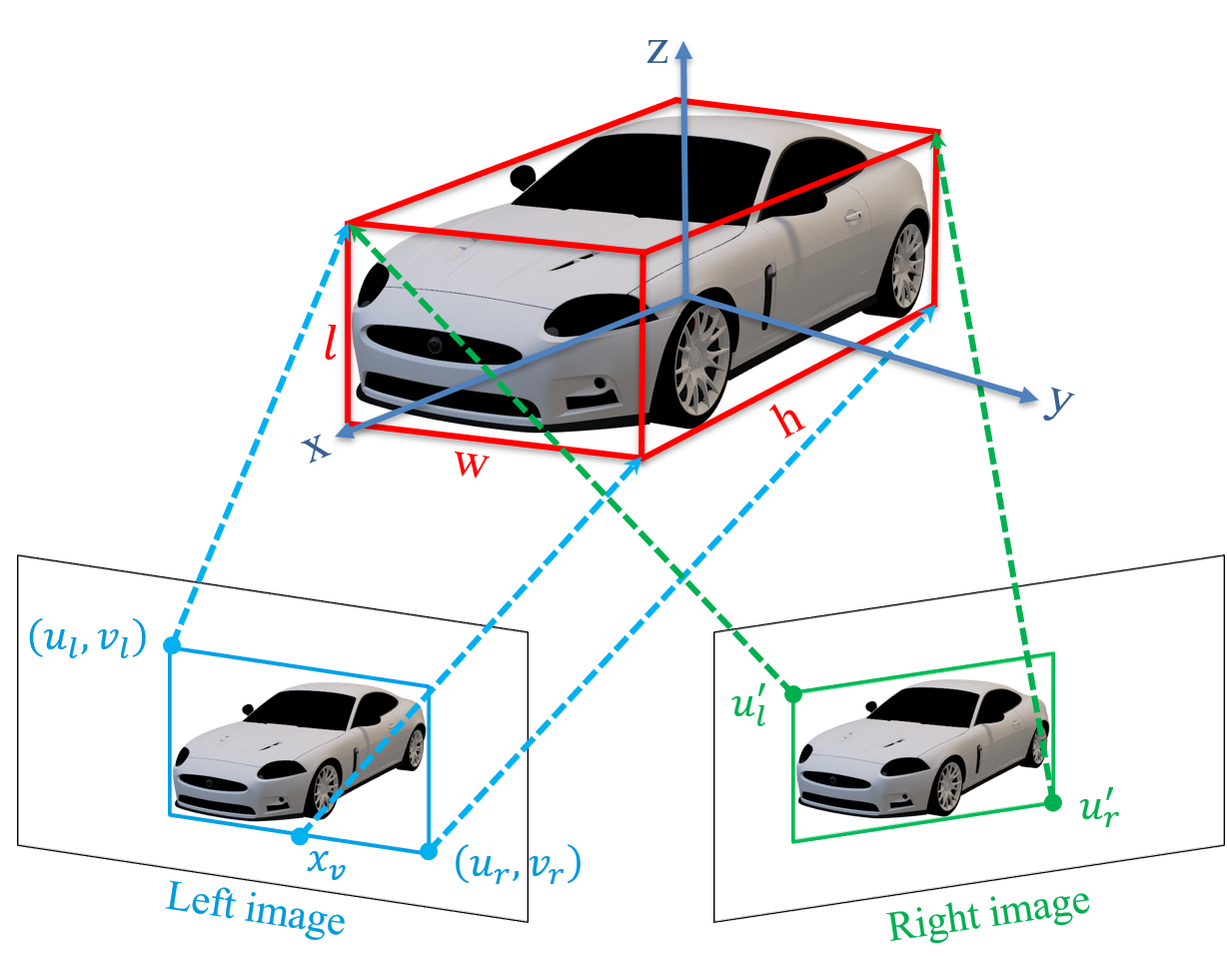}
	\end{center}
	\caption{3D box estimation with stereo key-point constraints.}
	\label{f}
\end{figure}

According to the predicted results in Sect. \ref{sosp}, seven key values 
$Z=\left \{\left ( u_{l}, v_{l} \right ) ,\left (u_{r}, v_{r}\right ), u_{l}^{\prime}, u_{r}^{\prime}, x_{v}\right \}$ represent the upper left and lower right coordinates of the 2D box of the left image, the left and right abscissas of the 2D box in the right image, and the abscissa of the perspective key vertex. As presented in the figure \ref{f}, according to the sparse geometric relationship between 2D and 3D, a set of constraint equations can be constructed:

\begin{small}
	\begin{eqnarray}
	\left\{\begin{array}{l}
	u_{l}=\left(x-\frac{w}{2} \cos \theta-\frac{l}{2} \sin \theta\right) /\left(z+\frac{w}{2} \sin \theta-\frac{l}{2} \cos \theta\right) \\
	v_{l}=\left(y-\frac{h}{2}\right) /\left(z+\frac{w}{2} \sin \theta-\frac{l}{2} \cos \theta\right) \\
	u_{r}=\left(x+\frac{w}{2} \cos \theta+\frac{l}{2} \sin \theta\right) /\left(z-\frac{w}{2} \sin \theta+\frac{l}{2} \cos \theta\right) \\
	v_{r}=\left(y+\frac{h}{2}\right) /\left(z-\frac{w}{2} \sin \theta+\frac{l}{2} \cos \theta\right) \\
	u_{l}^{\prime}=\left(x-b-\frac{w}{2} \cos \theta-\frac{l}{2} \sin \theta\right) /\left(z+\frac{w}{2} \sin \theta-\frac{l}{2} \cos \theta\right) \\
	u_{r}^{\prime}=\left(x-b+\frac{w}{2} \cos \theta+\frac{l}{2} \sin \theta\right) /\left(z-\frac{w}{2} \sin \theta+\frac{l}{2} \cos \theta\right) \\
	u_{p}=\left(x+\frac{w}{2} \cos \theta-\frac{l}{2} \sin \theta\right) /\left(z-\frac{w}{2} \sin \theta-\frac{l}{2} \cos \theta\right)
	\end{array}\right.
	\end{eqnarray}
\end{small}

Where $b$ represents the baseline length of the stereo cameras, $w$, $h$, and $l$ denote the regression size, and $x,y,z$ represent the coordinates of the 3D bounding box's center point. These multivariate equations were solved by the Gauss-Newton method.
\label{3D}

\begin{algorithm} 
	\label{alg}%
	\footnotesize
	\caption{3D Object Classification Strategy} 
	\LinesNumbered 
	\KwIn{Number $l$ of objects; The vertical coordinate of the upper left corner $y1$ and the lower right corner $y2$ of the box; A zero list $ depth\_line$ with a length of $1280$;
	}
	\KwOut{Id of the occluded object $o$ and the unoccluded object $u$}
	
	\For{$i=0;i\le l;i++$}{
		\emph{Read object depth coordinates $z$}\; 
		\For{$j=y1[i];j\le y2[i];j++$}{\label{forins}
			
			$pixel = depth\_line[j]$\;
			\If{$pixel= 0$ }
			{ 
				$depth\_line[j] = z$\;
			}
			
			\ElseIf{$z < depth\_line[j]:$}
			{ 
				$depth\_line[j] = (z + pixel) / 2 $\; 
	} 	}} 
	\For{$k=0;k\le l;k++$}{\label{forins}
		\emph{Read object depth coordinates $z$}\; 
		$left\_v = right\_v =True $\;
		\If{$depth\_line[y1[k]] < z$}{$left\_v = False$\;}
		\If{ $depth\_line[y2[k]] < z$}{	$right\_v = False$\;}
		\If{ $right\_v= False$ and $left\_v= False$}{$o\gets k $\; }	
		\Else{$u\gets k$\; }		
	}
			
\end{algorithm}

\begin{figure}[t]
	\begin{center}
		\includegraphics[width=8cm,height=9.5cm]{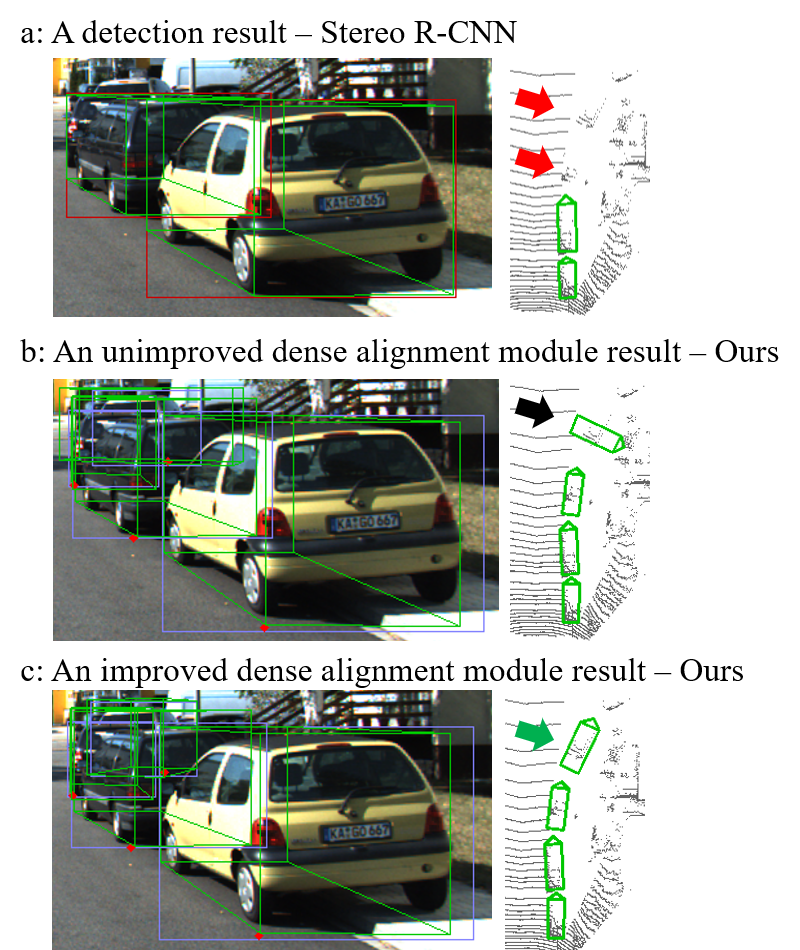}
		\scriptsize
	\end{center}
	\caption{Comparisons between Stereo CenterNet and Stereo R-CNN. In this figure, there are four objects with labels as car. The left image is the 3D box detection result and the right depicts the aerial view. The {\color{red}red arrows} represent the missed detection results, while the \textbf{black arrow} depicts the wrong detection results, and the {\color{green}green arrow} represents the true detection results.
	}
	\label{g}
\end{figure}

\noindent{\bf Dense Alignment} After obtaining the rough 3D bounding box in Sect \ref{3D}, to further optimize the position of the 3D box in space, we adopt the dense alignment module of Peiliang Li et al.\cite {Stereor-cnn}. Fig. \ref{g} presents the detection result on a scene in the KITTI dataset. As illustrated in Fig. \ref{g}(b), when the object is severely occluded, the dense alignment module exhibits the wrong result. However, as presented in Fig. \ref{g}(a), the same phenomenon does not emerge in Stereo R-CNN. After checking the code, we inferred that a possible reason for this phenomenon is that Stereo R-CNN discarded the severely occluded samples during training. Therefore, it is challenging to maintain the complete training samples and circumvent the negative optimization of the dense alignment module. 

To address this challenge, after the geometric construction module estimates the 3D bounding box, we classified these objects according to the detected 2D, 3D box, and depth information. For occluded objects, the geometric estimation results will be used directly. For objects that were not severely excluded, they will be put into the dense alignment module for optimization. After defining the optimization strategy, we implemented the classification procedure as Algorithm \ref{alg}. As illustrated in Fig. \ref{g}(c), for heavily occluded objects, the proposed SC exhibits correct optimization results, owing to our improved dense alignment module.

\begin{table*}[htbp]
	\centering
	
	\caption{Stereo average precision(AP) of 2D detection methods for car category, evaluated on the KITTI \textit{validation/test} set.}	
	\resizebox{140mm}{13mm}{ 	 
		\begin{tabular}{c|c|ccc|ccc|ccc|c}
			\hline
			\multirow{3}[6]{*}{Method} & \multirow{3}[6]{*}{Backbone} & \multicolumn{9}{c|}{IOU=0.7 [\textit{val / test}]   }                                    & \multirow{3}[6]{*}{Gap} \\
			\cline{3-11}      &       & \multicolumn{3}{c|}{Left} & \multicolumn{3}{c|}{Right} & \multicolumn{3}{c|}{ Stereo} &  \\
			\cline{3-11}      &       & Easy  & Moderate  & Hard  & Easy  & Moderate  & Hard  & Easy  & Moderate & Hard  &  \\
			\hline
			\hline
			CenterNet & DLA-34 & 97.1  & 87.9  & 79.3  & $-$     & $-$     & $-$     & $-$     & $-$     & $-$     & $-$ \\
			\hline
			Stereo R-CNN & ResNet-101 & 98.73/93.98 & 88.48/85.98 & 71.26/71.25  & 98.7  & 88.5  & 71.3  & \textbf{98.5} & 88.3  & 71.1  & -0.53 \\
			O-C Stereo & $-$     & \textBF{98.87}/87.39 & \textBF{90.53}/74.6 & \textBF{81.05}/62.56 & \textBF{98.9} & \textBF{90.5} & \textBF{80.9} & 98.4  & \textBF{90.4} & \textBF{80.7} & -0.92 \\
			\hline
			Ours  & ResNet-18 & 90.3  & 80    & 70.1  & 89.2  & 78.5  & 70.7  & 90    & 79.2  & 70.3  & -0.9 \\
			Ours  & DLA-34 & 97.94/\textBF{96.61} & 89.83/\textBF{91.27} & 80.83/\textBF{83.50}  & 97.6  & 88.4  & 79.8  & 97.9  & 89.4  & 80.5  & \textBF{-0.34} \\
			\hline
		\end{tabular}%
	}
	\label{ta}%
\end{table*}%
\subsection{Implementation Details}

\noindent{\bf Data Augementation.} We filled the original image $1280\times384$ for training and testing, then flipped the training set image horizontally, and swapped the images left and right to obtain a new stereo images; hence, our dataset was twice the original training number. The training data adopted the random scaling data enhancement method. Because the 3D information was inconsistent with the data enhancement, the proportional enhancement method was not suitable for the length, width, and height regressions.

\noindent{\bf Hyperparameter.} By default, we set the downsampling factor $R=4$. According to previous studies \cite{simonelli2019disentangling}, we adopted the average size of the car as $\left [\bar{L},\bar{W},\bar{H}\right ]^{\top }=\left [3.88,1.63,1.53\right ]^{\top }$. Set $\alpha =0.6$ in all the experiments. To prevent excessive vertex coordinates from overflowing the image during random scaling, we set the probability of random scaling to $0.5$. In the inference process, we set thresholds $0.25$ and $0.1$ to filter the main center and perspective key points of the left image, respectively.

\noindent{\bf Training.} We adopted the AdamW\cite{adamw} optimizer on a NVIDIA Tesla V100 GPU for 45 epochs of training. The initial learning rate was set to $1.5\times 10^{-4} $. The learning rate was reduced 10 times at the 40-th epoch. The backbone network was initialized by a classification model pre-trained on the ImageNet dataset. Regarding ResNet-18, we trained with 16 batch sizes for approximately 15 h. For DLA-34, we trained with 8 batch sizes for approximately 24 h.

\section{Experiments}

We evaluated the performance of our method on the popular KITTI 3D object detection dataset \cite{geiger2012we}, which contains 7481 training images and 7518 testing images. Based on 3DOP \cite{3DOP}, we split the training data into \textit{training} set(3712 images) and \textit{validation} set(3769 images). KITTI classifies objects into three levels: \textit{Easy}, \textit{Moderate} and \textit{Hard}, according to the occlusion/truncation and object size of each object category in the 2D image. The methods were further evaluated using different IoU criteria per class(IoU $\ge$ 0.7 for \textit{Car} and IoU $\ge$ 0.5 for \textit{Pedestrian} and \textit{Cyclist}). We mainly performed comparisons with other stereo images input methods in the car class, and a fully ablation study. We also benchmarked our results on the online KITTI \textit{test} server.

\begin{table*}[htbp]
	\centering
	\scriptsize 
	\caption{Car Localization and Detection. $AP_{BEV}/AP_{3D}$ on \textit{validation} set. The italicized data in the Runtime column indicates the data obtained from the original article, while the others represent the results of experiments performed on the same device. ``gap1'' indicates the gap between our results and the best results of other methods that require extra labels. ``gap2'' indicates the gap between our results and the best results of other methods that do not require extra labels.}
	\begin{tabular}{c|c|c||ccc|c}
		\hline
		\multirow{2}[4]{*}{Sensor} & \multirow{2}[4]{*}{Method} & \multirow{2}[4]{*}{Extra Labels} & \multicolumn{3}{c|}{$AP_{BEV}/AP_{3D}$(IOU=0.7)} & \multirow{2}[4]{*}{Runtime} \bigstrut\\
		\cline{4-6}      &       &       & Easy  & Moderate & Hard  &  \bigstrut\\
		\hline
		Mono  & M3D-RPN & No    & 25.94 / 20.27 & 21.18 / 17.06 & 17.90 / 15.21 & \textit{0.16s} \bigstrut\\
		\hline
		\hline
		Stereo & 3DOP  & Yes   & 12.63 / 6.55 & 9.49 / 5.07 & 7.59 / 4.10 & $-$ \bigstrut[t]\\
		Stereo & PL:F-PointNet & Yes   & 72.8 / 59.4 & 51.8 / 39.8 & 44.0 / 33.5 & \textit{0.08s} \\
		Stereo & PL:AVOD & Yes   & 74.9 / 61.9 & 56.8 / 45.3 & 49.0 / 39.0 & \textit{0.51s} \\
		Stereo & OC-Stereo & Yes   & 77.66 / 64.07 & 65.95 / 48.34 & 51.20 / 40.39 & \textit{0.35s} \\
		Stereo & ZoomNet & Yes   & 78.68 / 62.96 & \textBF{66.19} / \textBF{50.47} & \textBF{57.60} \textbf{/43.63} & $-$ \\
		Stereo & Disp R-CNN & Yes   & \textBF{77.63} / \textBF{64.29} & 64.38 / 47.73 & 50.68 / 40.11 & \textit{0.42s} \bigstrut[b]\\
		\hline
		\hline
		Stereo & TLNet & No    & \multicolumn{1}{l}{29.22 / 18.15} & \multicolumn{1}{l}{21.88 / 14.26} & \multicolumn{1}{l|}{18.83 / 13.72} & $-$ \bigstrut[t]\\
		Stereo & Stereo R-CNN & No    & \multicolumn{1}{l}{68.50 / 54.11} & \multicolumn{1}{l}{48.30 / 36.69} & \multicolumn{1}{l|}{41.47 / 31.07} & 0.2s \\
		Stereo & IDA-3D & No    & \multicolumn{1}{l}{70.68 / 54.97} & \multicolumn{1}{l}{50.21 / 37.45} & \multicolumn{1}{l|}{42.93 / 32.23} & \textit{0.08s} \bigstrut[b]\\
		\hline
		Stereo & Ours(ResNet-18) & No    & \multicolumn{1}{l}{59.34 / 43.67} & \multicolumn{1}{l}{43.99 / 32.71} & \multicolumn{1}{l|}{36.86 / 27.06} & \textBF{0.027s} \bigstrut[t]\\
		Stereo & Ours(DLA-34) & No    & \multicolumn{1}{l}{\textBF{71.26} / \textBF{55.25}} & \multicolumn{1}{l}{\textBF{53.27} / \textBF{41.44}} & \multicolumn{1}{l|}{\textBF{45.53} / \textBF{35.13}} & 0.043s \bigstrut[b]\\
		\hline
		\hline
		$-$     & gap1   & $-$     & -6.00 / -9.07  & -12.92 / -9.03     & -12.07 / -8.50& $-$ \bigstrut[t]\\
		$-$     & gap2   & $-$     & +0.58 / +0.28 & +3.06 / +3.99 & +2.60 / +2.90    & $-$ \bigstrut[b]\\
		\hline
	\end{tabular}%
	
	
	\label{tb}%
\end{table*}%

\subsection{Detection on KITTI}

\noindent{\bf 2D Detection Performance.} We compared the proposed SC with other stereo 3D detectors that contain left and right associate components, where the stereo $AP_{2D}$ metric is defined by Stereo R-CNN \cite{Stereor-cnn}. As presented in Table \ref{ta}, the proposed SC exhibits a higher detection precision on the left image than the CenterNet \cite{centernet} in \textit{Easy/Moderate/Hard} cases. In addition, the proposed SC correlated the left and right objects without extra computation. In the \textit{validation} set, the performance of our left-right objects correlation method outperforms or has comparable results to Stereo R-CNN, while the $AP_{2D}$ of the moderate and hard cases in the left image increased by \textbf{1.35\%} and \textbf{9.57\%}, respectively. 

It is important to note that it is difficult to accurately measure the performance of the box association method by solely reporting the stereo $AP_{2D}$. We observed that the Left $AP_{2D}$ is usually higher than the right and stereo $AP_{2D}$ in these methods; hence, we calculated the sum of the gap between the stereo and left $AP_{2D}$ in the three cases. As presented in Table \ref{ta}, the proposed box association method has the smallest gap. In the \textit{test} set, with DLA-34, the proposed SC obtains \textbf{96.61\%/91.27\%/83.50\%} and outperforms those of existing stereo SOTA studies. The detailed performance can be found online. Accurate stereo detection provides sufficient constraints for 3D box estimation.

\noindent{\bf 3D Detection Performance.} We compared SC with other stereo 3D detectors in Table \ref{tb}. In addition to the average precision for the aerial view($AP_{BEV}$) and 3D box($AP_{3D}$), we also provided comparison of inference time. Specifically, our method outperforms all monocular image based 3D detector. For the \textit{validation} set, the proposed SC surpasses all stereo methods, except those with extra labels. Among stereo 3D detectors without additional labels, using ResNet-18 backbone, the SC provides the best inference time and runs at 37 FPS. DLA34 runs at 23 FPS with \textbf{55.25\%/41.44\%/35.13\%}$AP_{3D}$. This is twice as fast as IDA-3D and reports \textbf{0.32\%/3.99\%/2.90\%}$AP_{3D}$, which is more accurate in the moderate case at 0.7 IoU.

As presented in Table \ref{tc}, the proposed SC exhibits better accuracy than IDA-3D \cite{IDA} on the KITTI \textit{test} set. The proposed SC adaptively switches between geometric constraints and subpixel disparity estimation to achieve better depth estimation for occluded and truncated objects. This explains why $AP_{BEV}$ and $AP_{3D}$ improved better in moderate and hard cases. We also compared the proposed SC with stereo detectors that contain extra labels. It is inappropriate to perform comparisons with these methods; hence, we only list them for reference. Compared with the Pseudo-LiDAR \cite{wang2019pseudo} using F-PointNet \cite{F-PointNet} as the detector, the proposed SC does not only achieve comparable detection performance, but also its inference time is only \textit{1/10}.

\begin{figure}[t]
	\begin{center}
		\includegraphics[width=6.7cm,height=5.5cm]{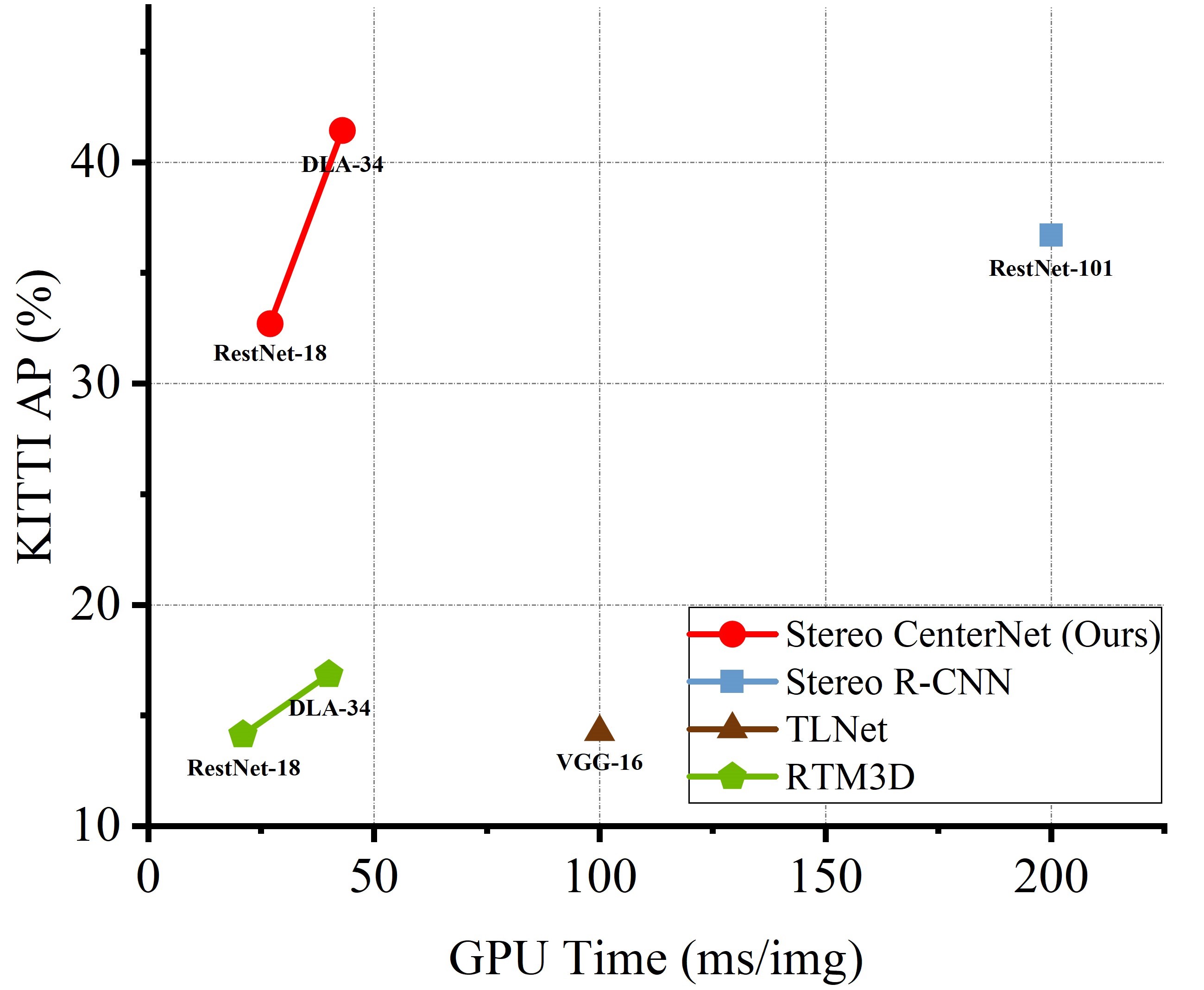}
	\end{center}
	\caption{Stereo 3D detection performance on KITTI (Moderate: IOU=0.7) validation set. The results of the comparison with baseline; the proposed SC achieves the best speed-accuracy trade-off.}
	\label{a}
\end{figure}

We also present the results of the comparison with baseline, as illustrated in Figure \ref{a}. We report the single-model speed-accuracy of the proposed SC with different backbones. The variant with DLA-34 outperforms the best stereo key-point 3D detector, Stereo R-CNN (41.4\% vs 36.6\%) while running more than 4$\times$ faster. The variant with the same backbones far outperforms the accurate monocular keypoint 3D detector, RTM3D (41.4\% $AP_{3D}$ in 43 ms vs 16.9\% $AP_{3D}$ in 38 ms), although it is only approximately 0.2$\times$ the increase in reasoning time.

\begin{table}[htbp]
	\centering
	\scriptsize 
		\caption{Car Localization and Detection. $AP_{BEV}/AP_{3D}$ on \textit{test} set. The italicizations in the method column indicates that the methods require extra labels, while the others are without extra labels.}
	\resizebox{85mm}{20mm}{ 
		\begin{tabular}{c|c|c|c|c|c|c|c}
			\hline
			\multirow{2}[4]{*}{Sensor} & \multirow{2}[4]{*}{Method} & \multicolumn{3}{c|}{3D Detection AP (\%)} & \multicolumn{3}{c}{BEV Detection AP (\%)} \bigstrut\\
			\cline{3-8}      &       & Easy  & \textBF{Moderate} & Hard  & Easy  & \textBF{Moderate} & Hard \bigstrut\\
			\hline
			\hline
			LiDAR & RT3D  & 23.74 & 19.14 & 18.86 & 56.44 & 44    & 42.34 \bigstrut\\
			\hline
			\hline
			Mono  & M3D-RPN & 14.76 & 9.71  & 	7.42 & 21.02 & 13.67 & 10.23 \bigstrut[t]\\
			Mono  & RTM3D & 14.41 & 10.34 & 8.77  & 19.17 & 14.2  & 11.99 \bigstrut[b]\\
			\hline
			\hline
			Stereo & \textit{RT3DStereo} & 29.9  & 23.28 & 18.96 & 58.81 & 46.82 & 38.38 \bigstrut[t]\\
			Stereo & \textit{PL:AVOD} & 54.53 & 34.05 & 28.25 & 67.3  & 45    & 38.4 \bigstrut[b]\\
			\hline
			\hline
			Stereo & TLNet & 7.64  & 4.37  & 3.74  & 13.71 & 7.69  & 6.73 \bigstrut[t]\\
			Stereo & IDA-3D & 45.09 & 29.32 & 23.13 & $-$     & $-$     & $-$ \\
			Stereo & Stereo R-CNN & 47.58 & 30.23 & 23.72  & 61.92 & 41.31 & 33.42 \\
			Stereo & SC(Ours) & 49.94 & 31.3  & 25.62 & 62.97 & 42.12 & 35.37 \bigstrut[b]\\
			\hline
		\end{tabular}%
		
	}	
	\label{tc}%
\end{table}%

\subsection{Ablation Study}
In this section, we present the results of the experiments conducted by comparing different left-right association, training, key point, and dense alignment strategies. These experiments are performed in the car category of \textit{training/validation} in the KITTI dataset. If not specified, the backbone for all experiments adopts the DLA-34 network.

\noindent{\bf Left-Right Association Strategy.} CenterNet \cite{centernet} adopts the center heatmap to detect objects; hence, it is difficult to associate left-right objects and circumvent additional computation. We tested two anchor-free associate strategies for two different Gaussian kernels: regress all detection heads on the right image (including rights center heatmap, rights center offset, right objects width and left-right distances) and solely regress right objects width and left-right distances. We experimented with different combinations to demonstrate their effects on detection performance, and the obtained results are presented in Table \ref{td}. The heatmap considering the aspect ratio improves the detection accuracy. No significant difference in accuracy exists between the two correlation methods, and the inference speed can be increased by 2 FPS by solely using the L-R distance. 

The result obtained from selecting all components is unsatisfactory. A possible reason is that the loss function contains several components from sub-branches; hence, the training procedure was complex and difficult to learn effective knowledge. Therefore, we solely predicted the right objects width and L-R distances to correlate the object, thereby reducing the amount of network calculations.
\label{Ablation}
\begin{table}[htbp]
	\centering
	\scriptsize
		\caption{Results are evaluated at a moderate level with IoU $=$ 0.7. Aspect Ratio stands for considering the aspect ratio of the heatmap in the kernel, and Right stands for the detection head containing the right object heatmap and center offset.}
	\begin{tabular}{c|c|c|c}
		\hline
		w/ Aspect Ratio & w/ Right & FPS   & $AP_{2D} / AP_{BEV} / AP_{3D}$ \bigstrut\\
		\hline
		$\times$ & $\surd$  & 21    & 83.45 / 46.94 / 37.87 \bigstrut[t]\\
		$\times$ & $\times$ & 23    & 88.65 / 50.68 / 38.44 \bigstrut[b]\\
		\hline
		$\surd$  & $\surd$  & 21    & \textBF{89.99} / 52.52 / 40.93 \bigstrut[t]\\
		$\surd$	 & $\times$ & 23    & 89.83 / \textBF{53.27} / \textBF{41.44} \bigstrut[b]\\
		\hline
	\end{tabular}%
	\label{td}%
\end{table}%

\begin{figure*}[t]
	\centering
	\includegraphics[width=15.12cm,height=4cm]{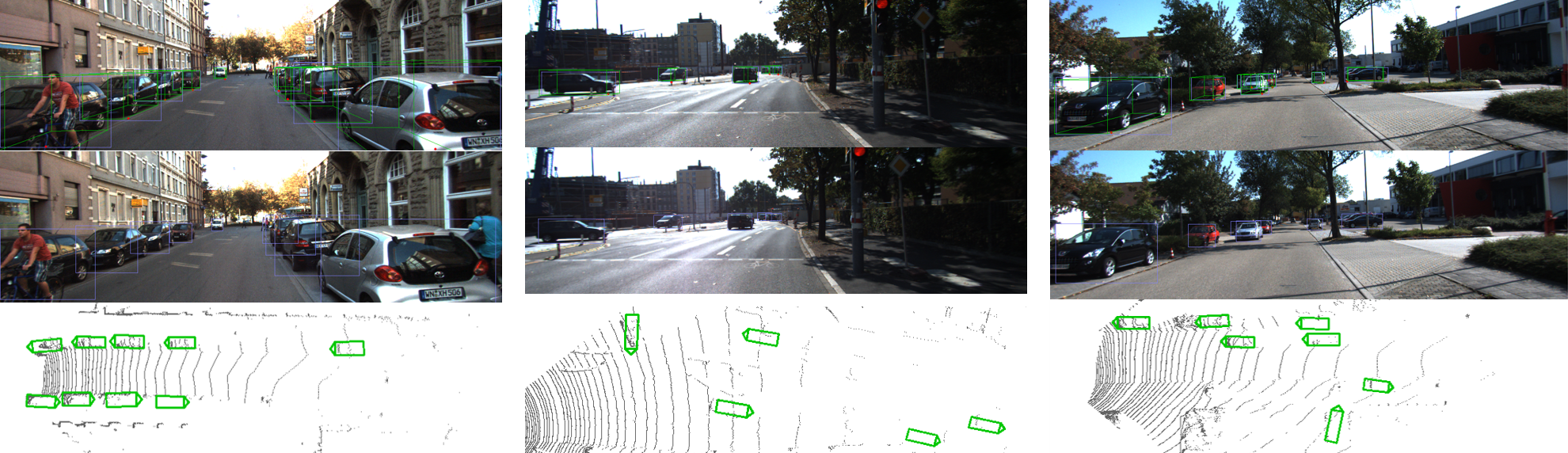}
	
	\caption{Quantitative results of multiple scenarios in the KITTI dataset. The first row presents the predicted 3D bounding boxes drawn from the detection results of the left image, the second row depicts the 2D bounding boxes in the right-eye image, and the third row presents the aerial view image.}
	\label{i}
\end{figure*}

\noindent{\bf Training Strategy.} During training, we adopted three strategies to enhance model performance: flip enhancement, uncertainty weight \cite{kendall2018multi}, and AdamW \cite{adamw} optimizer. Subsequently, we discussed their importance in our method, as presented in the Table \ref{te}. We conducted different combinations of experiments on the three strategies. Uncertain weight circumvents the manual adjustment of weights while achieving better accuracy. The stereo flip data enhancement doubles the number of training set and alleviates the imbalance between positive and negative samples. The AdamW optimizer improves accuracy (+0.57\% $AP_{3D}$) better than the Adam \cite{kingma2014adam} optimizer, and improves training speed. These strategies have played positive roles in ensuring that the proposed SC achieved better performance.
\begin{table}[htbp]
	\centering
	\scriptsize
		\caption{Evaluate the results at a moderate IoU level $=$ 0.7. ``Flip'' implies adopting random horizontal flipping augmentation. ``Uncert'' means using uncertain weight. ``AdamW'' means applying AdamW optimizer. All experiments are conducted in the same settings.}
	\begin{tabular}{cccc}
		\hline
		Flip  & Uncert & AdamW & $AP_{2D} / AP_{BEV} / AP_{3D}$ \bigstrut\\
		\hline
		&       &       & 79.10 / 42.97 / 30.20 \bigstrut[t]\\
		& $\surd$     &       & 81.24 / 42.72 / 32.88 \\
		$\surd$     &       &       & 89.80 / 52.23 / 40.86 \\
		$\surd$     & $\surd$     &       & 89.80 / 52.23 / 40.87 \\
		$\surd$     & $\surd$     & $\surd$     & \textBF{89.83} / \textBF{53.27} / \textBF{41.44} \bigstrut[b]\\
		\hline
	\end{tabular}%
	\label{te}%
\end{table}%

\noindent{\bf Keypoint Strategy.} We tested three detection methods for the key points at the bottom of the 3D box: direct regression to the center point depth, direct detection of the perspective key points, and detection and classification of the four vertices at the bottom. We compared $AP_{3D}$ in the two IOU criteria. As presented in Table \ref{tf}, we performed experiments on the three strategies. Four-point detection can provide higher accuracy, and is more conducive to the construction of 3D box. In addition to the perspective key points, the points can also provide pixel-level constraints for the dense alignment module, which is more appropriate for the optimization of the 3D box position.
\begin{table}[htbp]
	\centering
	\scriptsize	
	\caption{Detect different numbers of key points in our geometric construction module. ``Without key points'' means direct regression to the center point depth.}
	\begin{tabular}{c|ccc|ccc}
		\hline
		\multirow{2}[4]{*}{method} & \multicolumn{3}{c|}{$AP_{3D}$(IOU=0.5)} & \multicolumn{3}{c}{$ AP_{3D}$(IOU=0.7)} \bigstrut\\
		\cline{2-7}      & Easy  & \textBF{Moderate} & Hard  & Easy  & \textBF{Moderate} & Hard \bigstrut\\
		\hline
		w/o kp & 79.85 & 63.12 & 55.08 & 41.11 & 30.22 & 25.19 \bigstrut[t]\\
		w/ 1kp & 86.41 & 73.57 & 59.16 & 52.10  & 39.01 & 32.76 \\
		w/ 4kp & \textBF{86.54} & \textBF{73.98} & \textBF{65.70}  & \textBF{55.25} & \textBF{41.44} & \textBF{35.13} \bigstrut[b]\\
		\hline
	\end{tabular}%
	\label{tf}%
\end{table}%

\noindent{\bf Dense Alignment Strategy.} In this part of the experiment, we present the advantages of our dense alignment module strategy. We conducted experiments in three cases: no dense alignment, dense alignment, and our dense alignment strategy. As shown in the Table \ref{tg}, our dense alignment module strategy exhibits the highest accuracy and a better accuracy improvement on hard samples.
\begin{table}[htbp]
	\centering
	\scriptsize
	\caption{Evaluations of no-dense, dense and our dense alignment modules in the KITTI validation set. We use ResNet-18 as the backbone.}
	\resizebox{85mm}{9mm}{ 
		\begin{tabular}{c|ccc|ccc}
			\hline
			\multirow{2}[4]{*}{Method} & \multicolumn{3}{c|}{$AP_{BEV}$(IOU=0.7)} & \multicolumn{3}{c}{$AP_{3D}$(IOU=0.7)} \bigstrut\\
			\cline{2-7}      & Easy  & \textBF{Moderate} & Hard  & Easy  & \textBF{Moderate} & Hard \bigstrut\\
			\hline
			w/o Alignment & 13.41 & 11.28 & 10.34 & 8.65  & 7.21  & 6.16 \bigstrut[t]\\
			w/ Alignment & 59.21 & 43.87 & 36.59 & 43.54 & 32.59 & 26.95 \\
			w/ Alignment(Ours) & \textBF{59.34} & \textBF{43.99} & \textBF{36.86} & \textBF{43.67} & \textBF{32.71} & \textBF{27.06} \bigstrut[b]\\
			\hline
		\end{tabular}%
	}
	\label{tg}%
\end{table}%

\noindent{\bf Pedestrian and Cyclist detection.} In the KITTI object detection benchmark, the training samples of \textit{Pedestrian} and \textit{Cyclist} are limited; hence, it is more difficult than detecting \textit{car} category. Because most image-based methods do not exhibit the evaluation results of \textit{Pedestrian} and \textit{Cyclist}, we solely report the available results of the original paper. We present the pedestrian and cyclist detection results on KITTI \textit{validation} set in Table \ref{tpc}. In fact, PL:F-PointNet and DSGN \cite{Dsgn} are both methods that utilize extra labels. The proposed SC exhibits more accurate results than these methos on $AP_{2D}$, but worse results on $AP_{BEV}$ and $AP_{3D}$.

\begin{table}[htbp]
	\centering
	\scriptsize 
	\caption{Comparison of results for Pedestrian and Cyclist on KITTI \textit{validation} set.}
	\resizebox{85mm}{18mm}{ 
		\begin{tabular}{c|c|c|c}
			\hline
			\multirow{2}[4]{*}{Method} & $AP_{2D}$  & $AP_{BEV}$  & $AP_{3D}$ \bigstrut\\
			\cline{2-4}      & E / M / H & E / M / H & E / M / H \bigstrut\\
			\hline
			\multicolumn{4}{c}{\textit{Pedestrian}} \bigstrut\\
			\hline
			PL: F-PointNet & $-$      & 41.30 / 34.90 / 30.10 & 33.80 / 27.40 / 24.00 \bigstrut[t]\\
			DSGN  & 59.06 / 54.00 / 49.65 & \textBF{47.92} / \textBF{41.15} / \textBF{36.08} &  \textBF{40.16} / \textBF{33.85} / \textBF{29.43} \\
			SC(Ours) & \textBF{68.17} / \textBF{59.59} / \textBF{51.25} & 29.21 / 26.41 / 21.87 & 27.57 / 24.71 / 20.73 \bigstrut[b]\\
			\hline
			\multicolumn{4}{c}{\textit{Cyclist}} \bigstrut\\
			\hline
			PL: F-PointNet & $-$      & \textBF{47.60} / \textBF{29.90} / \textBF{27.00} & \textBF{41.30} / \textBF{25.20} / \textBF{24.90} \bigstrut[t]\\
			DSGN  & 49.38 / 33.97 / 32.40 &  41.86 / 25.98 / 24.87 & 37.87 / 24.27 / 23.15 \\
			SC(Ours) & \textBF{76.16} / \textBF{51.10} / \textBF{50.39} & 37.45 / 24.83 / 23.99 & 36.59 / 24.10 / 23.37 \bigstrut[b]\\
			\hline
		\end{tabular}%
	}	
	\label{tpc}%
\end{table}%

\subsection{Qualitative Results}
We present the qualitative results of a number of scenarios in the KITTI dataset in the Figure \ref {i}. We present the corresponding stereo box, 3D box, and aerial view on the left and right images. It can be observed that in general street scenes, the proposed SC can accurately detect vehicles in the scene, and the detected 3D frame can be optimally aligned with the LiDAR point cloud. It also detected a few small objects that were occluded and far away.

\section{Conclusion and Future Work}
This research proposed SC, a faster 3D object detection method for stereo images. We addressed the 3D detection problem as a key point detection problem via a combination of learning and geometry. The proposed SC achieved the best speed-accuracy trade-off without requiring anchor-based 2D detection methods, depth estimation, and LiDAR monitoring. We infer that the image-based methods have substantial potential in the 3D field. However, the proposed framework only learned a few right-images information to facilitate geometric calculations, and the stereo feature network was not carefully designed to be slightly time-consuming. Therefore, we can attempt to further refine and simplify the framework by learning stereo information from single images while ensuring detection performance. Another main limitation of our method is that SC has not been tested in other scenarios except for the autonomous driving scenario; hence, the performance of the application in other scenarios remains unknown. In the future, we will test the application and deployment of this method in other scenarios such as indoor simultaneous localization and mapping (SLAM) \cite{xw} and remote surgery \cite{DBLP1,DBLP2,DBLP3}.

\section*{Acknowledgment}
The authors acknowledge financial support from the National Natural Science Foundation of China under Grant No.61772068, the Finance science and technology project of Hainan province (No. ZDYF2019009), the China Postdoctoral Science Foundation under Grant No.2020M680352, the Guangdong Basic and Applied Basic Research Foundation under Grant No.2020A1515110463, the Scientific and Technological Innovation Foundation of Shunde Graduate School, USTB under Grant No.2020BH011. The computing work is partly supported by USTB MatCom of Beijing Advanced Innovation Center for Materials Genome Engineering.

\bibliographystyle{unsrt}
\bibliography{new}

\begin{thebibliography}{10}

\bibitem{li2018stereo}
Peiliang Li, Tong Qin, et~al.
\newblock Stereo vision-based semantic 3d object and ego-motion tracking for
  autonomous driving.
\newblock In {\em Proceedings of the European Conference on Computer Vision
  (ECCV)}, pages 646--661, 2018.

\bibitem{cubeslam}
Shichao Yang and Sebastian Scherer.
\newblock Cubeslam: Monocular 3-d object slam.
\newblock {\em IEEE Transactions on Robotics}, 35(4):925--938, 2019.

\bibitem{Pv-rcnn}
Shaoshuai Shi, Chaoxu Guo, Li~Jiang, Zhe Wang, Jianping Shi, Xiaogang Wang, and
  Hongsheng Li.
\newblock Pv-rcnn: Point-voxel feature set abstraction for 3d object detection.
\newblock In {\em Proceedings of the IEEE/CVF Conference on Computer Vision and
  Pattern Recognition}, pages 10529--10538, 2020.

\bibitem{3dssd}
Zetong Yang, Yanan Sun, Shu Liu, and Jiaya Jia.
\newblock 3dssd: Point-based 3d single stage object detector.
\newblock In {\em Proceedings of the IEEE/CVF Conference on Computer Vision and
  Pattern Recognition}, pages 11040--11048, 2020.

\bibitem{chen2017multi}
Xiaozhi Chen, Huimin Ma, Ji~Wan, Bo~Li, and Tian Xia.
\newblock Multi-view 3d object detection network for autonomous driving.
\newblock In {\em Proceedings of the IEEE Conference on Computer Vision and
  Pattern Recognition}, pages 1907--1915, 2017.

\bibitem{Std}
Zetong Yang, Yanan Sun, Shu Liu, Xiaoyong Shen, and Jiaya Jia.
\newblock Std: Sparse-to-dense 3d object detector for point cloud.
\newblock In {\em Proceedings of the IEEE International Conference on Computer
  Vision}, pages 1951--1960, 2019.

\bibitem{ZoomNet}
Zhenbo Xu, Wei Zhang, Xiaoqing Ye, Xiao Tan, Wei Yang, Shilei Wen, Errui Ding,
  Ajin Meng, and Liusheng Huang.
\newblock Zoomnet: Part-aware adaptive zooming neural network for 3d object
  detection.
\newblock In {\em AAAI}, pages 12557--12564, 2020.

\bibitem{stereo-matching}
Daniel Scharstein and Richard Szeliski.
\newblock A taxonomy and evaluation of dense two-frame stereo correspondence
  algorithms.
\newblock {\em Int. J. Comput. Vis.}, 47(1-3):7--42, 2002.

\bibitem{epnp}
Vincent Lepetit, Francesc Moreno{-}Noguer, and Pascal Fua.
\newblock Ep\emph{n}p: An accurate \emph{O}(\emph{n}) solution to the
  p\emph{n}p problem.
\newblock {\em Int. J. Comput. Vis.}, 81(2):155--166, 2009.

\bibitem{ObjectNet3D}
Yu~Xiang, Wonhui Kim, Wei Chen, Jingwei Ji, Christopher~B. Choy, Hao Su,
  Roozbeh Mottaghi, Leonidas~J. Guibas, and Silvio Savarese.
\newblock Objectnet3d: {A} large scale database for 3d object recognition.
\newblock In Bastian Leibe, Jiri Matas, Nicu Sebe, and Max Welling, editors,
  {\em Computer Vision - {ECCV} 2016 - 14th European Conference, Amsterdam, The
  Netherlands, October 11-14, 2016, Proceedings, Part {VIII}}, volume 9912 of
  {\em Lecture Notes in Computer Science}, pages 160--176. Springer, 2016.

\bibitem{geiger2012we}
Andreas Geiger, Philip Lenz, and Raquel Urtasun.
\newblock Are we ready for autonomous driving? the kitti vision benchmark
  suite.
\newblock In {\em 2012 IEEE Conference on Computer Vision and Pattern
  Recognition}, pages 3354--3361. IEEE, 2012.

\bibitem{ApolloScape}
Xinyu Huang, Xinjing Cheng, Qichuan Geng, Binbin Cao, Dingfu Zhou, Peng Wang,
  Yuanqing Lin, and Ruigang Yang.
\newblock The apolloscape dataset for autonomous driving.
\newblock In {\em 2018 {IEEE} Conference on Computer Vision and Pattern
  Recognition Workshops, {CVPR} Workshops 2018, Salt Lake City, UT, USA, June
  18-22, 2018}, pages 954--960. Computer Vision Foundation / {IEEE} Computer
  Society, 2018.

\bibitem{nuScenes}
Holger Caesar, Varun Bankiti, Alex~H. Lang, Sourabh Vora, Venice~Erin Liong,
  Qiang Xu, Anush Krishnan, Yu~Pan, Giancarlo Baldan, and Oscar Beijbom.
\newblock nuscenes: {A} multimodal dataset for autonomous driving.
\newblock In {\em 2020 {IEEE/CVF} Conference on Computer Vision and Pattern
  Recognition, {CVPR} 2020, Seattle, WA, USA, June 13-19, 2020}, pages
  11618--11628. Computer Vision Foundation / {IEEE}, 2020.

\bibitem{SVBS}
Peiliang Li, Tong Qin, and Shaojie Shen.
\newblock Stereo vision-based semantic 3d object and ego-motion tracking for
  autonomous driving.
\newblock In Vittorio Ferrari, Martial Hebert, Cristian Sminchisescu, and Yair
  Weiss, editors, {\em Computer Vision - {ECCV} 2018 - 15th European
  Conference, Munich, Germany, September 8-14, 2018, Proceedings, Part {II}},
  volume 11206 of {\em Lecture Notes in Computer Science}, pages 664--679.
  Springer, 2018.

\bibitem{Structured}
Shichao Yang and Sebastian~A. Scherer.
\newblock Monocular object and plane {SLAM} in structured environments.
\newblock {\em {IEEE} Robotics Autom. Lett.}, 4(4):3145--3152, 2019.

\bibitem{Stereor-cnn}
Peiliang Li, Xiaozhi Chen, and Shaojie Shen.
\newblock Stereo r-cnn based 3d object detection for autonomous driving.
\newblock In {\em Proceedings of the IEEE Conference on Computer Vision and
  Pattern Recognition}, pages 7644--7652, 2019.

\bibitem{wang2019pseudo}
Yan Wang, Wei-Lun Chao, Divyansh Garg, Bharath Hariharan, Mark Campbell, and
  Kilian~Q Weinberger.
\newblock Pseudo-lidar from visual depth estimation: Bridging the gap in 3d
  object detection for autonomous driving.
\newblock In {\em Proceedings of the IEEE Conference on Computer Vision and
  Pattern Recognition}, pages 8445--8453, 2019.

\bibitem{IDA}
Wanli Peng, Hao Pan, He~Liu, and Yi~Sun.
\newblock Ida-3d: Instance-depth-aware 3d object detection from stereo vision
  for autonomous driving.
\newblock In {\em Proceedings of the IEEE/CVF Conference on Computer Vision and
  Pattern Recognition}, pages 13015--13024, 2020.

\bibitem{Dsgn}
Yilun Chen, Shu Liu, Xiaoyong Shen, and Jiaya Jia.
\newblock Dsgn: Deep stereo geometry network for 3d object detection.
\newblock In {\em Proceedings of the IEEE/CVF Conference on Computer Vision and
  Pattern Recognition}, pages 12536--12545, 2020.

\bibitem{Disp}
Jiaming Sun, Linghao Chen, Yiming Xie, Siyu Zhang, Qinhong Jiang, Xiaowei Zhou,
  and Hujun Bao.
\newblock Disp r-cnn: Stereo 3d object detection via shape prior guided
  instance disparity estimation.
\newblock In {\em Proceedings of the IEEE/CVF Conference on Computer Vision and
  Pattern Recognition}, pages 10548--10557, 2020.

\bibitem{3DOP}
Xiaozhi Chen, Kaustav Kundu, Yukun Zhu, Huimin Ma, Sanja Fidler, and Raquel
  Urtasun.
\newblock 3d object proposals using stereo imagery for accurate object class
  detection.
\newblock {\em IEEE transactions on pattern analysis and machine intelligence},
  40(5):1259--1272, 2017.

\bibitem{maskrcnn}
Kaiming He, Georgia Gkioxari, Piotr Doll{\'{a}}r, and Ross~B. Girshick.
\newblock Mask {R-CNN}.
\newblock In {\em {IEEE} International Conference on Computer Vision, {ICCV}
  2017, Venice, Italy, October 22-29, 2017}, pages 2980--2988. {IEEE} Computer
  Society, 2017.

\bibitem{tian2019fcos}
Zhi Tian, Chunhua Shen, Hao Chen, and Tong He.
\newblock Fcos: Fully convolutional one-stage object detection.
\newblock In {\em Proceedings of the IEEE/CVF International Conference on
  Computer Vision}, pages 9627--9636, 2019.

\bibitem{li2019gs3d}
Buyu Li, Wanli Ouyang, Lu~Sheng, Xingyu Zeng, and Xiaogang Wang.
\newblock Gs3d: An efficient 3d object detection framework for autonomous
  driving.
\newblock In {\em Proceedings of the IEEE Conference on Computer Vision and
  Pattern Recognition}, pages 1019--1028, 2019.

\bibitem{shi2020distance}
Xuepeng Shi, Zhixiang Chen, and Tae-Kyun Kim.
\newblock Distance-normalized unified representation for monocular 3d object
  detection.
\newblock In {\em European Conference on Computer Vision}, pages 91--107.
  Springer, 2020.

\bibitem{RTM3D}
Peixuan Li, Huaici Zhao, Pengfei Liu, and Feidao Cao.
\newblock Rtm3d: Real-time monocular 3d detection from object keypoints for
  autonomous driving.
\newblock {\em arXiv preprint arXiv:2001.03343}, 2020.

\bibitem{chen2020monopair}
Yongjian Chen, Lei Tai, Kai Sun, and Mingyang Li.
\newblock Monopair: Monocular 3d object detection using pairwise spatial
  relationships.
\newblock In {\em Proceedings of the IEEE/CVF Conference on Computer Vision and
  Pattern Recognition}, pages 12093--12102, 2020.

\bibitem{SMOKE}
Zechen Liu, Zizhang Wu, and Roland T{\'o}th.
\newblock Smoke: Single-stage monocular 3d object detection via keypoint
  estimation.
\newblock In {\em Proceedings of the IEEE/CVF Conference on Computer Vision and
  Pattern Recognition Workshops}, pages 996--997, 2020.

\bibitem{beker2020monocular}
Deniz Beker, Hiroharu Kato, Mihai~Adrian Morariu, Takahiro Ando, Toru Matsuoka,
  Wadim Kehl, and Adrien Gaidon.
\newblock Monocular differentiable rendering for self-supervised 3d object
  detection.
\newblock {\em arXiv preprint arXiv:2009.14524}, 2020.

\bibitem{qin2019triangulation}
Zengyi Qin, Jinglu Wang, and Yan Lu.
\newblock Triangulation learning network: from monocular to stereo 3d object
  detection.
\newblock In {\em Proceedings of the IEEE/CVF Conference on Computer Vision and
  Pattern Recognition}, pages 7615--7623, 2019.

\bibitem{you2019pseudo}
Yurong You, Yan Wang, Wei-Lun Chao, Divyansh Garg, Geoff Pleiss, Bharath
  Hariharan, Mark Campbell, and Kilian~Q Weinberger.
\newblock Pseudo-lidar++: Accurate depth for 3d object detection in autonomous
  driving.
\newblock {\em arXiv preprint arXiv:1906.06310}, 2019.

\bibitem{qian2020end}
Rui Qian, Divyansh Garg, Yan Wang, Yurong You, Serge Belongie, Bharath
  Hariharan, Mark Campbell, Kilian~Q Weinberger, and Wei-Lun Chao.
\newblock End-to-end pseudo-lidar for image-based 3d object detection.
\newblock In {\em Proceedings of the IEEE/CVF Conference on Computer Vision and
  Pattern Recognition}, pages 5881--5890, 2020.

\bibitem{pon2020object}
Alex~D Pon, Jason Ku, Chengyao Li, and Steven~L Waslander.
\newblock Object-centric stereo matching for 3d object detection.
\newblock In {\em 2020 IEEE International Conference on Robotics and Automation
  (ICRA)}, pages 8383--8389. IEEE, 2020.

\bibitem{centernet}
Xingyi Zhou, Dequan Wang, and Philipp Kr{\"a}henb{\"u}hl.
\newblock Objects as points.
\newblock {\em arXiv preprint arXiv:1904.07850}, 2019.

\bibitem{resnet}
Kaiming He, Xiangyu Zhang, Shaoqing Ren, and Jian Sun.
\newblock Deep residual learning for image recognition.
\newblock In {\em Proceedings of the IEEE conference on computer vision and
  pattern recognition}, pages 770--778, 2016.

\bibitem{dla}
Fisher Yu, Dequan Wang, Evan Shelhamer, and Trevor Darrell.
\newblock Deep layer aggregation.
\newblock In {\em Proceedings of the IEEE conference on computer vision and
  pattern recognition}, pages 2403--2412, 2018.

\bibitem{dcnv2}
Xizhou Zhu, Han Hu, Stephen Lin, and Jifeng Dai.
\newblock Deformable convnets v2: More deformable, better results.
\newblock In {\em Proceedings of the IEEE Conference on Computer Vision and
  Pattern Recognition}, pages 9308--9316, 2019.

\bibitem{ttf}
Zili Liu, Tu~Zheng, Guodong Xu, Zheng Yang, Haifeng Liu, and Deng Cai.
\newblock Training-time-friendly network for real-time object detection.
\newblock In {\em AAAI}, pages 11685--11692, 2020.

\bibitem{law2018cornernet}
Hei Law and Jia Deng.
\newblock Cornernet: Detecting objects as paired keypoints.
\newblock In {\em Proceedings of the European Conference on Computer Vision
  (ECCV)}, pages 734--750, 2018.

\bibitem{focalloss}
Tsung-Yi Lin, Priya Goyal, Ross Girshick, Kaiming He, and Piotr Doll{\'a}r.
\newblock Focal loss for dense object detection.
\newblock In {\em Proceedings of the IEEE international conference on computer
  vision}, pages 2980--2988, 2017.

\bibitem{eigen2014depth}
David Eigen, Christian Puhrsch, and Rob Fergus.
\newblock Depth map prediction from a single image using a multi-scale deep
  network.
\newblock In {\em Advances in neural information processing systems}, pages
  2366--2374, 2014.

\bibitem{mousavian20173d}
Arsalan Mousavian, Dragomir Anguelov, John Flynn, and Jana Kosecka.
\newblock 3d bounding box estimation using deep learning and geometry.
\newblock In {\em Proceedings of the IEEE Conference on Computer Vision and
  Pattern Recognition}, pages 7074--7082, 2017.

\bibitem{kendall2018multi}
Alex Kendall, Yarin Gal, and Roberto Cipolla.
\newblock Multi-task learning using uncertainty to weigh losses for scene
  geometry and semantics.
\newblock In {\em Proceedings of the IEEE conference on computer vision and
  pattern recognition}, pages 7482--7491, 2018.

\bibitem{simonelli2019disentangling}
Andrea Simonelli, Samuel~Rota Bulo, Lorenzo Porzi, Manuel L{\'o}pez-Antequera,
  and Peter Kontschieder.
\newblock Disentangling monocular 3d object detection.
\newblock In {\em Proceedings of the IEEE International Conference on Computer
  Vision}, pages 1991--1999, 2019.

\bibitem{adamw}
Ilya Loshchilov and Frank Hutter.
\newblock Fixing weight decay regularization in adam.
\newblock 2018.

\bibitem{F-PointNet}
Charles~R. Qi, Wei Liu, Chenxia Wu, Hao Su, and Leonidas~J. Guibas.
\newblock Frustum pointnets for 3d object detection from {RGB-D} data.
\newblock In {\em 2018 {IEEE} Conference on Computer Vision and Pattern
  Recognition, {CVPR} 2018, Salt Lake City, UT, USA, June 18-22, 2018}, pages
  918--927. Computer Vision Foundation / {IEEE} Computer Society, 2018.

\bibitem{kingma2014adam}
Diederik~P Kingma and Jimmy Ba.
\newblock Adam: A method for stochastic optimization.
\newblock {\em arXiv preprint arXiv:1412.6980}, 2014.

\bibitem{xw}
Yanmin Wu, Yunzhou Zhang, Delong Zhu, Yonghui Feng, Sonya Coleman, and Dermot
  Kerr.
\newblock {EAO-SLAM:} monocular semi-dense object {SLAM} based on ensemble data
  association.
\newblock In {\em {IEEE/RSJ} International Conference on Intelligent Robots and
  Systems, {IROS} 2020, Las Vegas, NV, USA, October 24, 2020 - January 24,
  2021}, pages 4966--4973. {IEEE}, 2020.

\bibitem{DBLP1}
Hang Su, Andrea Mariani, Salih~Ertug Ovur, Arianna Menciassi, Giancarlo
  Ferrigno, and Elena~De Momi.
\newblock Toward teaching by demonstration for robot-assisted minimally
  invasive surgery.
\newblock {\em {IEEE} Trans Autom. Sci. Eng.}, 18(2):484--494, 2021.

\bibitem{DBLP2}
Hang Su, Wen Qi, Chenguang Yang, Juan Sebasti{\'{a}}n~Sandoval Ar{\'{e}}valo,
  Giancarlo Ferrigno, and Elena~De Momi.
\newblock Deep neural network approach in robot tool dynamics identification
  for bilateral teleoperation.
\newblock {\em {IEEE} Robotics Autom. Lett.}, 5(2):2943--2949, 2020.

\bibitem{DBLP3}
Hang Su, Wen Qi, Yingbai Hu, Hamid~Reza Karimi, Giancarlo Ferrigno, and Elena
  De~Momi.
\newblock An incremental learning framework for human-like redundancy
  optimization of anthropomorphic manipulators.
\newblock {\em IEEE Transactions on Industrial Informatics}, 2020.

\end{thebibliography}
\end{document}